\documentclass[letterpaper]{article} 
\usepackage{aaai2027}  
\usepackage[hyphens]{url}  
\usepackage{graphicx} 
\usepackage{natbib}  
\usepackage{caption} 
\usepackage{algorithm}
\usepackage{algorithmic}

\usepackage{newfloat}
\usepackage{listings}
\usepackage{amsmath}

\DeclareCaptionStyle{ruled}{labelfont=normalfont,labelsep=colon,strut=off} 
\floatstyle{ruled}
\newfloat{listing}{tb}{lst}{}
\floatname{listing}{Listing}

\usepackage{booktabs}
\usepackage{xcolor}
\usepackage{pifont}
\newcommand{\cmark}{\ding{51}}
\newcommand{\xmark}{\ding{55}}
\usepackage{amssymb}
\title{SPARE: Structural Parameter-Free Affinity Regularization for Flow Matching}
\author{
    Zong-Wei Hong, 
    Jinglun Li,
    Shen Zhang,
    Yuhan Liu,
    Linze Li\textsuperscript{$\ddagger$},
    Yao Tang\textsuperscript{$\dagger$}
}
\affiliations{
    JIIOV Technology\\
    {\tt\small \{zongwei.hong, jinglun.li, shen.zhang, yuhan.liu, linze.li, yao.tang\}@jiiov.com}
}

\begin{document}
\nocopyright
\maketitle

\makeatletter
\def\blfootnote{\gdef\@thefnmark{}\@footnotetext}
\makeatother
\blfootnote{\textsuperscript{$\dagger$} Corresponding author. \textsuperscript{$\ddagger$} Project leader.}

\begin{abstract}
Denoising diffusion transformers achieve strong generation quality but converge slowly during training. Regularizing their internal representations has emerged as an effective accelerator, yet existing methods split into two families with complementary costs. Target-based methods strengthen representations by aligning them to external features, which requires an external encoder and a learnable projection head to bridge feature spaces. Target-free methods hold no reference at all, and can only repel the model's own features across samples or layers, discarding whatever structure the data contains. Prior work suggests that spatial structure, rather than global semantics, drives the gains of alignment. We therefore ask whether such structure can serve as a target directly, and whether it exists not only within an image but across images. Our key insight is that the clean data latent already carries this structure in the relations among its tokens, where a relation is the similarity between two tokens, a single scalar comparable across feature spaces without a projection head. We propose Structural Parameter-free Affinity Regularization (SPARE), a regularizer that matches the pairwise affinities of intermediate tokens to those of the clean latents. To exploit this structure fully, SPARE extends the matching to token pairs across images, precisely the pairs that prior target-free methods repel by default, and calibrates both relation types with a single learning objective. On ImageNet $256 \times  256$ with SiT backbones under matched 400K-iteration budgets, SPARE adds no encoder, head, or parameters and only 0.08 GB of training memory, yet attains the lowest FID among parameter-free regularizers in every tested setting, recovers 37 to 54\% of REPA's FID reduction, and improves over REPA when combined with it, reaching FID 1.90 under classifier-free guidance at 1M iterations. Ablations show that, under identical machinery, matching cross-image affinities improves generation while repelling them degrades it.
\end{abstract}

\begin{figure}[!t]
\centering
\includegraphics[width=0.9\columnwidth]{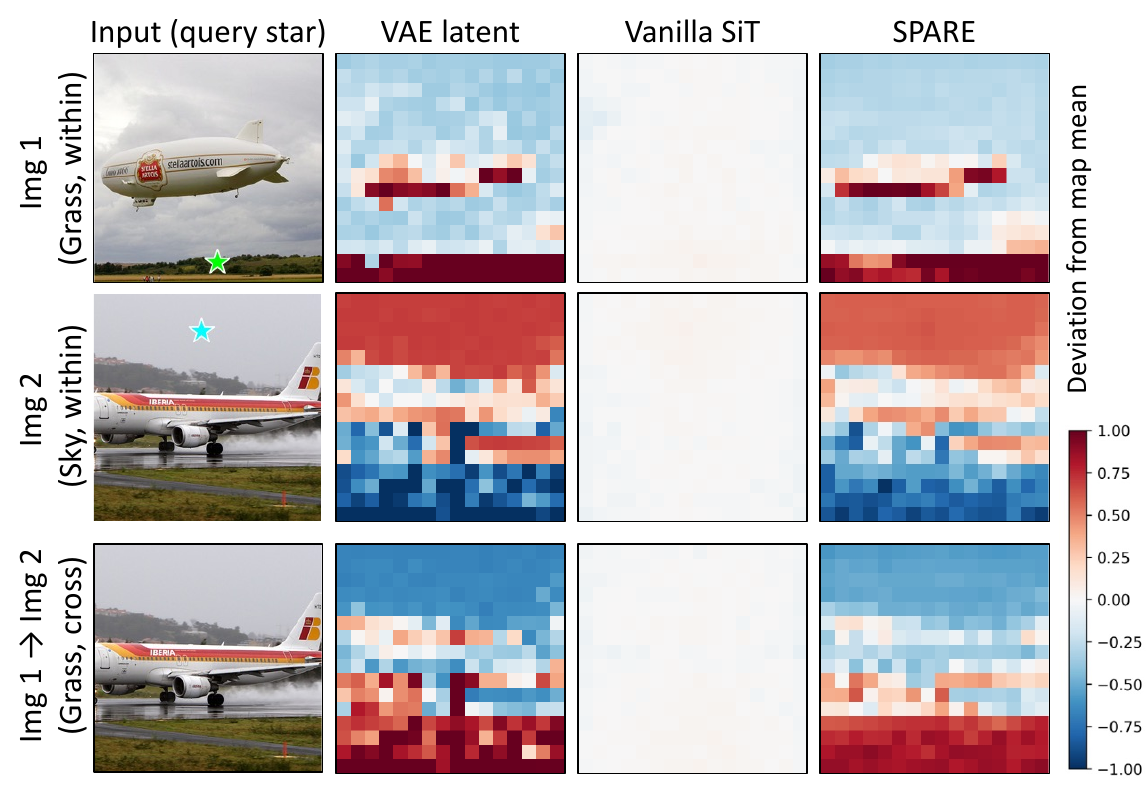}
\caption{Each map shows the cosine similarity between the query token (star) and all tokens, plotted as deviation from the map's mean. Model features are taken from layer 3 of SiT-XL/2 at 400K iterations with noised input at timestep 0.5, while the VAE latent is computed from the clean image. The clean latent carries two kinds of affinity structure. Within a single image, affinities follow object boundaries (rows 1–2, one query per image). Across images in a batch, the grass query of image 1 localizes the grass region of image 2 (row 3). Vanilla training recovers neither (mean cosine 0.95, near-uniform maps). SPARE recovers both directly from the data, with no projection head and no added parameters, and improves generation quality.}
\label{fig:affi}
\end{figure}

\begin{figure*}[t]
\centering
\includegraphics[width=\textwidth]{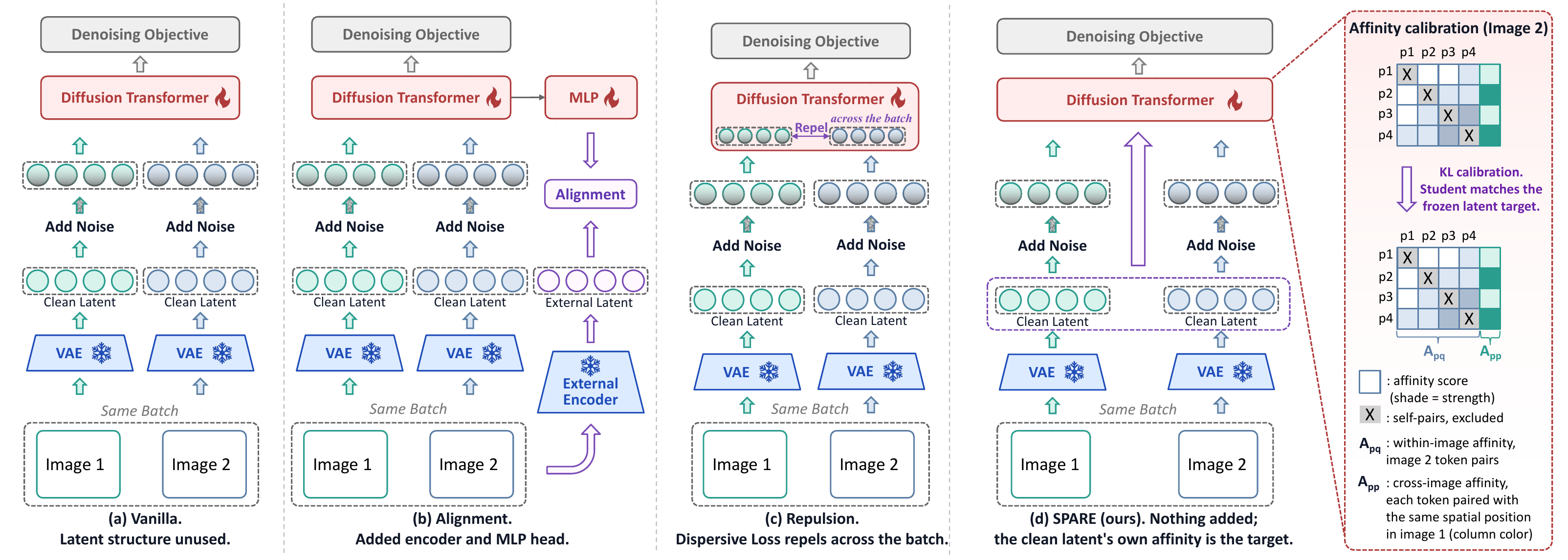}
\caption{\textbf{Four ways to regularize a diffusion transformer's
representations.} (a)~Vanilla training, where the latent's structure
goes unused. (b)~Target-based alignment (REPA, SRA, SRA2) matches
each image's tokens to reference features through a learnable MLP
head (red); the reference may come from a frozen external encoder
(blue, drawn) or from internal substitutes. (c)~Target-free
objectives drop the target and repel the model's own features
uniformly; Dispersive Loss (drawn) repels across the batch, and
DiverseDiT applies the same principle across blocks.
(d)~\textbf{SPARE (ours)} adds nothing. Pairwise affinities of
intermediate tokens are calibrated to those of the clean latent,
across all positions within each image and at the same position
across images, so the data decides which relations should be strong
and which weak. The zoom-in, drawn for $|\mathcal{B}| = 2$ images
and $P = 4$ tokens, shows one image's affinity rows with self-pairs
excluded, split into within-image pairs ($A_{pq}$) and same-position
cross-image pairs ($A_{pp}$), and the KL calibration toward the
frozen latent target.}
\label{fig:teaser}
\end{figure*}

\section{Introduction}

Diffusion and flow-matching models~\cite{ho2020denoising,
song2020score, lipman2022flow, liu2022rectifiedflow,
albergo2025stochastic} have become the dominant paradigm for visual
generation. These models learn to reverse a gradual noising process,
or equivalently a velocity field that transports noise to data, and
achieve remarkable fidelity and diversity across image synthesis,
video generation, and controllable content
creation~\cite{dhariwal2021diffusion, rombach2022high,
ho2022video, zhang2023adding}. Scaling these models to transformer backbones~\cite{peebles2023scalable, ma2024sit} pushes quality further still, but training remains slow and compute-hungry. How to make such training more efficient has therefore become a central question.

One answer has proven remarkably effective. Diffusion transformers
generate better when their internal representations are better, and
two lines of work pursue this through added regularization.
\emph{Target-based} (Fig.~\ref{fig:teaser}b) alignment attracts intermediate representations toward an assigned reference. REPA~\cite{yu2024representation}
 draws the reference from a pretrained visual
encoder, tying representation
learning to models that are themselves massively expensive to
obtain. SRA~\cite{jiang2025sra} and SRA2~\cite{wang2026sra}
remove the external encoder, aligning instead to an EMA teacher and
to clean data features respectively, yet every member of this line
retains a learnable MLP projection head that
bridges the model's features into the reference space, and with it
extra trainable parameters serving only the auxiliary loss.
\emph{Target-free} diversity (Fig.~\ref{fig:teaser}c) drops both.
Rather than aligning representations with anything,
Dispersive~Loss~\cite{wang2025diffuse} simply repels samples within a
batch, and DiverseDiT~\cite{yang2026diversedit} applies the same
principle across transformer blocks, encouraging dispersion without reference to
what the data says.

Viewed together, these methods form a sequence of removals. Removing the encoder leaves the projection head in place. Removing the head discards the alignment target as well, and only repulsion remains. Whether the two must be discarded together has not been examined. iREPA~\cite{singh2025matters} reports that REPA's gains derive largely from the spatial structure of the target, the pairwise relations among its patch tokens, rather than from its global semantics. sREPA~\cite{xu2026srepa} confirms that supervising these relations is beneficial, yet it still draws them from an external encoder through a projection. A pairwise relation, however, requires neither. It is a single scalar similarity between two tokens, comparable across feature spaces without a projection and indifferent to which image either token resides in. We therefore ask whether the clean data latent itself carries such structure, within an image and across a batch. Fig.~\ref{fig:affi} answers both. Within an image, token affinities follow object boundaries. Across images, affinities at corresponding positions connect batch samples with consistent content, structure that repulsion alone cannot represent. Standard training imposes no such constraint, and the structure does not emerge on its own. The intermediate tokens of a vanilla SiT collapse toward mutual similarity, and their affinity maps reflect neither boundaries nor cross-image correspondences.

We instantiate this as Structural Parameter-free Affinity Regularization (\textbf{SPARE}, Fig.~\ref{fig:teaser}d), a regularizer that \emph{calibrates} the pairwise affinities of intermediate tokens, within each image and across the batch, to those of the clean data, via a single Kullback--Leibler (KL) divergence between joint affinity distributions. SPARE imports no external reference; the clean latent itself specifies which relations should be strong and which weak. On ImageNet $256\times256$ with SiT backbones at matched 400K-step budgets, SPARE attains the lowest FID among parameter-free regularizers in every tested setting, and recovers a substantial fraction of REPA's reduction. SPARE also composes with encoder-based alignment. Added to REPA's recipe, it improves SiT-XL/2 FID at 400K without guidance and at 1M with guidance, indicating that the two shape
complementary aspects of the representation.
\begin{itemize}
    \item We identify the clean data latent as a free, structured alignment target. Its token relations, single similarity scalars comparable across feature spaces, follow object boundaries within images and link content across them, yet vanilla training acquires neither.
    \item We propose SPARE, a parameter-free regularizer that calibrates the pairwise affinities of intermediate tokens to those of the clean latent through a single KL objective, with no encoder, no projection head, and no added parameters. To our knowledge, SPARE is the first to treat cross-image relations as a signal to calibrate rather than a default to repel.
    \item On ImageNet $256\times256$ under matched budgets, SPARE attains the lowest FID among parameter-free regularizers in every tested setting, recovers 37 to 54\% of REPA's gain with 0.08\,GB extra memory, and improves REPA to FID 1.90 at 1M iterations. Under identical machinery, matching cross-image affinities improves generation while repelling them degrades it.
\end{itemize}

 
\section{Related Work}
\textbf{Representation regularization for diffusion training.}
REPA~\cite{yu2024representation} introduced the recipe, a per-token
alignment loss at an intermediate block against
DINOv2~\cite{oquab2023dinov2} features through a learnable MLP head,
and follow-ups extend it rather than shrink it, unlocking joint VAE
tuning~\cite{leng2025repa}, entangling latents with a class
token~\cite{wu2025reg}, or moving the intervention into the
tokenizer itself~\cite{yao2025reconstruction, zheng2025diffusion}.
SRA~\cite{jiang2025sra} and SRA2~\cite{wang2026sra} remove the
external encoder, distilling against an EMA teacher and clean data
features respectively, while the head and the per-token loss carry
over unchanged. AHPA~\cite{min2026ahpa} deepens this encoder-free
line, drawing hierarchical targets from intermediate VAE features
and scheduling their mix with a timestep-conditioned router, at the
price of a projection head and a routing network. Dispersive
Loss~\cite{wang2025diffuse} abandons targets altogether, repelling
the model's own features within a batch, and
DiverseDiT~\cite{yang2026diversedit} extends the repulsion across
blocks through added skip connections and projections, so trainable
weights persist even without a target. LSEP~\cite{yun2025lsep}
replaces feature targets with class labels, coupling an
intermediate block to a jointly optimized linear probe, so the
signal is global and still flows through a learnable head. The overhead shrinks along this line of work, yet no method offers a structured target at zero added cost.

Closest to our motivation, iREPA~\cite{singh2025matters} perturbs
REPA's target and finds the gains track its spatial structure
rather than its global semantics. The evidence is observational.
The direct intervention is to supervise the relations themselves.
sREPA~\cite{xu2026srepa} does so inside the encoder-based recipe,
matching pairwise similarities against the DINOv2 teacher, and its
gains corroborate the structural reading. Yet the relations are
computed after the learnable projection, the encoder stays in the
loop, and every pair lives within one image. SPARE moves the intervention to the clean latent, where relations need neither an encoder nor a projection, calibrates the cross-image pairs that prior target-free methods repel by default, and leaves the backbone's training cost essentially unchanged.

\textbf{Relational objectives beyond diffusion.}
Supervising relations in place of individual features has precedent, though at a coarser granularity. In knowledge distillation, RKD~\cite{park2019relational} transfers the geometry of a teacher's embedding space through pairwise distances and angles among sample embeddings, and similarity-preserving distillation~\cite{tung2019similarity} matches the batch-level Gram matrix of activations, so inputs the teacher deems alike stay alike for the student. The motivation is familiar, relations survive changes of dimension, so the student need not share the teacher's width. ReSSL~\cite{zheng2021ressl} carries the idea to self-supervised learning, aligning similarity distributions between augmented views rather than forcing exact feature agreement. Across these settings, relational objectives serve discriminative ends and relate whole samples, one embedding to another. SPARE differs in granularity and in provenance. Its relations live among tokens, fine enough to encode object boundaries within an image and same-position correspondence across images, and its target is read off the clean data rather than a trained teacher.


\section{Method}

\subsection{Preliminaries}
\textbf{Problem Setup.}
Flow matching learns a vector field that transports a prior $p_0$ (a standard
Gaussian) to a target $p_1$. Following the latent-diffusion
framework~\cite{rombach2022high}, we operate in latent space, where each image
$I$ is encoded by a frozen, pretrained VAE into a clean
latent $\mathbf{x} \in \mathbb{R}^{C \times H \times W}$, upon which the flow
field is learned. Computed once per training sample at negligible cost, this
clean latent already encodes rich spatial structure, which we
later reuse as the reference for our regularizer.

\textbf{Stochastic Interpolants.}
The transport from noise to data is a time-dependent process described by a
stochastic interpolant~\cite{ma2024sit, albergo2025stochastic}: for $\epsilon \sim \mathcal{N}(0, I)$,
\begin{equation}
    \label{eq:interpolation}
    \mathbf{x}_t = \alpha_t \mathbf{x} + \sigma_t \epsilon, \qquad t \in [0, 1],
\end{equation}
with time-dependent schedules $\alpha_t, \sigma_t$ subject to
$\alpha_1 = \sigma_0 = 1$ and $\alpha_0 = \sigma_1 = 0$. Although non-linear
parameterizations are possible, linear schedules suffice in practice, so we set
$\alpha_t = t$ and $\sigma_t = 1 - t$.

\textbf{Flow-Matching Objective.}
A velocity field $v_\theta$, parameterized by a diffusion transformer, transports samples along this path. Under the linear schedule the target
velocity is $\mathbf{x} - \epsilon$, and the model is trained by regressing onto
it:
\begin{equation}
    \label{eq:fm_loss}
    \mathcal{L}_{\text{FM}} = \mathbb{E}_{t, \mathbf{x}, \epsilon}
    \left[\, \big\| v_\theta(\mathbf{x}_t, t) - (\mathbf{x} - \epsilon) \big\|_2^2 \,\right].
\end{equation}
Beyond this output, the transformer produces intermediate token features
$\mathbf{h}^{(\ell)} \in \mathbb{R}^{P \times C_s}$ at each block $\ell$, where
$P$ is the number of spatial tokens and $C_s$ the hidden dimension.

\textbf{A Unified View of Representation Regularizers.}
Despite differing motivations, existing methods regularize the
intermediate representation through a shared template. Writing
$\mathbf{h}^{(\ell)}_{i} \in \mathbb{R}^{P \times C_s}$ for the token
matrix of the $i$-th sample in a batch $\mathcal{B}$ at layer $\ell$,
every method maps it by an operator $\phi$ and compares the result
against an assigned reference $\mathbf{r}_{i}$ under a discrepancy
$\mathcal{D}$,
\begin{equation}
    \label{eq:unified_reg}
    \mathcal{L}_{\text{reg}} = \mathbb{E}_{t,\mathbf{x},\epsilon}
    \Big[\, \frac{1}{|\mathcal{B}|}\sum_{i\in\mathcal{B}}
    \mathcal{D}\big(\, \phi(\mathbf{h}^{(\ell)}_{i}),\; \mathbf{r}_{i} \,\big) \,\Big],
\end{equation}
and the families differ only in the choice of $\phi$ and the
assignment of $\mathbf{r}_{i}$. \emph{Target-based} alignment (REPA,
SRA, SRA2) takes $\phi$ as a learnable MLP head and
$\mathbf{r}_{i}$ as the position-wise features of an external
reference. \emph{Target-free} diversity (Dispersive~Loss,
DiverseDiT) takes $\phi$ as the identity and $\mathbf{r}_{i}$ as the
model's own features, drawn from batch-mates or other blocks, with
$\mathcal{D}$ repelling rather than
attracting.

Within Eq.~\eqref{eq:unified_reg}, holding a target thus costs a
head, and dropping the head costs the target. The coupling follows
from a single premise, that $\phi$ outputs \emph{features}, and
features are comparable only within a shared space. An external
reference lives outside the model's space, so attraction requires a
learnable bridge. Remove the bridge, and the reference pool shrinks
to the model's own features, all computed from \emph{noised} inputs.
Head-free attraction is therefore confined to noised views, which
Dispersive Loss evaluated and found unprofitable, keeping repulsion
alone, a choice that presumes batch-mates share no structure worth
keeping. The clean latent is thus left supervising the output through
the position-wise regression of Eq.~\eqref{eq:fm_loss}, its spatial
structure given no head-free path to the representation.

SPARE opens this path with a single operator applied at two ranges.
Affinities of the clean latent describe which positions within an
image share a common structure and which positions across images
depict consistent content, precisely what uniform repulsion erases.
We align the model's affinities to the within-image target first,
then extend it across the batch, and the two terms form one
regularizer added to the flow-matching objective.

\begin{figure}[t]
\centering
\includegraphics[width=0.7\linewidth]{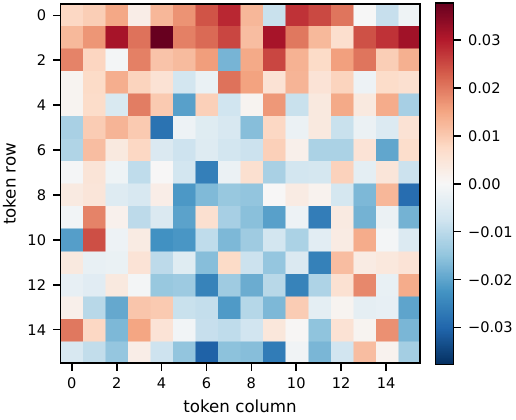}
\caption{\textbf{Cross-image same-position affinity of clean
latents.} For each spatial position, we average the cosine
similarity between the clean VAE latents of two \emph{different}
images at that same position (later formalized as
$A_{pp}(\mathbf{v}_i, \mathbf{v}_{i'})$), over ${\sim}20$K images in
class-shuffled batches. Values are shown as deviation from the
whole-map average (white $=$ map mean, red above, blue below). The
deviations are spatially organized, elevated across the top rows and
depressed over the middle and lower regions, matching typical scene
layout.}
\label{fig:motivation}
\end{figure}

\subsection{Within-Image Affinity as a Free Target}
\label{sec:intra}
For token matrices $\mathbf{u}, \mathbf{u}' \in \mathbb{R}^{P \times d}$ of
arbitrary width $d$, let $\hat{\mathbf{u}}^{\,p}$ denote the
$\ell_2$-normalized $p$-th row and define
\begin{equation}
  A_{pq}(\mathbf{u}, \mathbf{u}') \;=\;
  \bigl\langle \hat{\mathbf{u}}^{\,p}, \hat{\mathbf{u}}'^{\,q} \bigr\rangle,
  \label{eq:affinity}
\end{equation}
the cosine affinity between position $p$ of one matrix and position
$q$ of the other. We write
$A(\mathbf{u}, \mathbf{u}') \in \mathbb{R}^{P \times P}$ for the
matrix with entries $A_{pq}$, and abbreviate the self-affinity
$A(\mathbf{u}) := A(\mathbf{u}, \mathbf{u})$.
The map collapses the ambient dimension. Whatever the width, the
output lives in the same $P \times P$ space, so the dimension gap
that pushes prior work toward learnable heads simply disappears.

The target follows immediately, as the structure is already in the
pipeline. The clean latent is a spatially faithful encoding of the
image. Positions covering a common structure stay similar in latent
space, while unrelated regions do not, so the relations among its
positions mirror the spatial organization of the scene, before any
noise and without any external model. Concretely, given the clean
latent $\mathbf{x}_i \in \mathbb{R}^{C \times H \times W}$ of the
$i$-th image, we arrange it as $\mathbf{v}_i \in \mathbb{R}^{P \times C}$,
one row per token position, so that $\mathbf{v}_i$ and the token
matrix $\mathbf{h}^{(\ell)}_i \in \mathbb{R}^{P \times C_s}$ index
the same spatial positions. When the latent grid is finer than the
token grid, local pooling suffices. Then $A(\mathbf{v}_i)$ captures
which positions belong to a common structure, and
$A(\mathbf{h}^{(\ell)}_i)$ is directly comparable to it although
$C_s \neq C$.

In the vocabulary of Eq.~\eqref{eq:unified_reg},
$\phi = A(\cdot)$ is parameter-free,
$\mathbf{r}_i = A(\mathbf{v}_i)$ comes free, and $\mathcal{D}$
remains attractive, the combination both branches forgo. External
semantics is not thereby dispensable, and encoder-based methods
retain an edge in our experiments, but the structural component is
now priced at zero.

\subsection{Cross-Image Affinity as a Free Extension}
\label{sec:cross_image}
The target above draws on one image at a time, yet nothing in
Eq.~\eqref{eq:affinity} restricts its two arguments to the same
image, and the batch supplies $|\mathcal{B}|-1$ candidates for free.
Whether their affinities truly carry no structure worth keeping, as
uniform repulsion presumes, is testable. Natural photographs share
positional regularities in scene layout, so if any cross-image
structure survives, it should surface between \emph{different} images
at the \emph{same} position.

We test this directly. For each position $p$ we measure the mean of
$A_{pp}(\mathbf{v}_i, \mathbf{v}_{i'})$ over image pairs $i \neq i'$
from ${\sim}20$K images in class-shuffled batches
(Fig.~\ref{fig:motivation}). The per-position means are far from
uniform, reaching $1.5\times$ the map average across the top rows and
falling to $0.6\times$ over the middle and lower regions, a layout
that matches typical scene composition. Uniform repulsion would erase
all of it, the high affinities along with the low.
The probe fixes the design. We extend the target with same-position
pairs only, scored by $A_{pp}(\cdot,\cdot)$ on the latents and the
token matrices alike. Supervision touches matched positions alone,
yet the trained model recovers the full cross-image maps of
Fig.~\ref{fig:affi}, so the broader correspondence emerges rather
than being imposed.

\subsection{Training Objective}
\label{sec:objective}
The intra- and cross-image targets combine into a single
distribution-matching loss. Fix a layer $\ell$ and abbreviate
$\mathbf{h}_i = \mathbf{h}^{(\ell)}_i$. Treating each token as an
anchor, the candidate set of image $i \in \mathcal{B}$ at position
$p$ is
\begin{equation}
  \mathcal{C}(i,p) \;=\;
  \{(i,q) : q \neq p\} \,\cup\, \{(i',p) : i' \neq i\},
  \label{eq:candidates}
\end{equation}
the $P{-}1$ remaining positions of the same image (intra) and the
same position in the $|\mathcal{B}|{-}1$ other images (inter); the
anchor itself is excluded, its affinity being identically one. The
model scores every candidate with Eq.~\eqref{eq:affinity},
\begin{equation}
  s_{i,p}(c) =
  \begin{cases}
    A_{pq}(\mathbf{h}_i, \mathbf{h}_i), & c = (i,q), \\[2pt]
    A_{pp}(\mathbf{h}_i, \mathrm{sg}[\mathbf{h}_{i'}]), & c = (i',p),
  \end{cases}
  \label{eq:student_scores}
\end{equation}
where $\mathrm{sg}[\cdot]$ denotes stop-gradient. Within an image
both tokens receive gradient, while across images the candidate acts
as a fixed key, so no gradient couples the samples of a batch.
Collecting the scores over all candidates yields a vector
$s_{i,p} \in \mathbb{R}^{|\mathcal{C}(i,p)|}$ per anchor. The target
vector $t_{i,p}$ is built identically from the clean latents,
replacing $\mathbf{h}$ with $\mathbf{v}$ in
Eq.~\eqref{eq:student_scores}, and is detached throughout. A softmax
at temperature $\tau$, written $\sigma_\tau$, turns each vector into
a distribution over the candidate set, and the loss matches the two
distributions anchor-wise,
\begin{equation}
  \mathcal{L}_{\mathrm{SPARE}}
  = \frac{1}{|\mathcal{B}|\,P} \sum_{i,\,p}
  D_{\mathrm{KL}}\bigl(
    \sigma_\tau(t_{i,p}) \,\big\|\, \sigma_\tau(s_{i,p})
  \bigr),
  \label{eq:spare_loss}
\end{equation}
with the target as reference distribution. The final loss is
$\mathcal{L} = \mathcal{L}_{\mathrm{FM}} +
\lambda\,\mathcal{L}_{\mathrm{SPARE}}$, where $\lambda$ is a
scalar loss weight.
\section{Experiments}

\subsection{Experimental Setup}
\textbf{Implementation.}
All models train from scratch on class-conditional
ImageNet-1k~\cite{deng2009imagenet} $256\times256$ under the
original SiT recipe~\cite{ma2024sit} for 400K steps (AdamW,
learning rate $1\times10^{-4}$, batch size 256, EMA decay 0.9999,
seed 42), the default unless stated otherwise. Clean latents come
from the frozen \textit{sd-vae-ft-ema} tokenizer~\cite{rombach2022high},
precomputed before training and matched to the token grid by
$2\times2$ average pooling. SPARE alone supervises the output of
block $\ell=3$ with $\lambda=1$ and $\tau=0.5$ over $t\in[0,1]$ for
both backbones. For text-to-image generation we train
MMDiT~\cite{esser2024scaling} on MS-COCO~\cite{lin2014microsoft}
$256\times256$ under REPA's recipe (150K iterations, batch size
256, hidden dim.\ 768, depth 24), adding only the affinity loss.
In the REPA compositions, on ImageNet and MS-COCO alike, the
affinity loss attaches to block $\ell=2$, with all other settings
unchanged. We report FID~\cite{heusel2017gans},
IS~\cite{salimans2016improved}, sFID~\cite{nash2021generating},
Precision, and Recall~\cite{kynkaanniemi2019improved} on 50K
samples with the ADM suite~\cite{dhariwal2021diffusion}, under SDE
Euler--Maruyama and ODE Heun samplers with 250 steps. For T2I we
report FID on the validation split under the SDE sampler with 50
steps at guidance scale $w=2.0$.

\begin{table}[!t]
\centering
{\small
\setlength{\tabcolsep}{2pt}
\renewcommand{\arraystretch}{1.15}
\begin{tabular}{@{}lccccccc@{}}
\toprule
\textbf{Method} & \textbf{Enc.} & \textbf{Par.} & \textbf{FID}$\downarrow$ & \textbf{IS}$\uparrow$ & \textbf{sFID}$\downarrow$ & \textbf{Prec.}$\uparrow$ & \textbf{Rec.}$\uparrow$ \\
\midrule
\multicolumn{8}{@{}l}{\emph{SiT-B/2, SDE (E--M)}} \\
Baseline & \xmark & \xmark & 36.80 & 40.09 & 6.77 & 0.51 & 0.63 \\
+ REPA & \cmark & \cmark & 24.40 & 59.90 & 6.40 & 0.59 & 0.65 \\
+ DiverseDiT & \xmark & \cmark & \underline{28.05} & \underline{50.66} & \underline{6.04} & \underline{0.57} & 0.63 \\
+ SRA2 & \xmark & \cmark & 28.90 & -- & -- & -- & -- \\
+ Dispersive Loss$^\ddagger$ & \xmark & \xmark & 31.44 & 47.40 & \textbf{6.23} & 0.55 & 0.64 \\
+ \textbf{SPARE (Ours)} & \xmark & \xmark & \textbf{30.12} & \textbf{49.26} & 6.29 & \textbf{0.56} & \underline{\textbf{0.65}} \\
\addlinespace
\multicolumn{8}{@{}l}{\emph{SiT-B/2, ODE (Heun)}} \\
Baseline & \xmark & \xmark & 36.49 & 40.60 & 6.47 & 0.51 & 0.65 \\
+ Dispersive Loss$^\ddagger$ & \xmark & \xmark & 33.03 & 44.40 & \textbf{6.40} & 0.53 & \textbf{0.65} \\
+ \textbf{SPARE (Ours)} & \xmark & \xmark & \textbf{31.59} & \textbf{46.58} & 6.42 & \textbf{0.54} & 0.64 \\
\midrule
\multicolumn{8}{@{}l}{\emph{SiT-XL/2, SDE (E--M)}} \\
Baseline & \xmark & \xmark & 17.43 & 76.00 & 5.11 & 0.64 & 0.64 \\
+ REPA & \cmark & \cmark & 7.90 & 122.6 & 5.06 & 0.70 & 0.65 \\
+ DiverseDiT & \xmark & \cmark & 12.42 & \underline{95.01} & \underline{4.85} & \underline{0.68} & 0.63 \\
+ SRA2 & \xmark & \cmark & \underline{11.70} & -- & -- & -- & -- \\
+ Dispersive Loss$^\ddagger$ & \xmark & \xmark & 15.54 & 81.40 & 5.12 & \textbf{0.67} & 0.63 \\
+ \textbf{SPARE (Ours)} & \xmark & \xmark & \textbf{13.86} & \textbf{91.22} & \textbf{5.05} & \textbf{0.67} & \underline{\textbf{0.64}} \\
\addlinespace
\multicolumn{8}{@{}l}{\emph{SiT-XL/2, ODE (Heun)}} \\
Baseline & \xmark & \xmark & 18.12 & 72.43 & 5.11 & 0.63 & 0.64 \\
+ Dispersive Loss$^\ddagger$ & \xmark & \xmark & 16.73 & 76.00 & 5.08 & 0.64 & \textbf{0.64} \\
+ \textbf{SPARE (Ours)} & \xmark & \xmark & \textbf{14.34} & \textbf{86.20} & \textbf{5.05} & \textbf{0.66} & \textbf{0.64} \\
\addlinespace
\multicolumn{8}{@{}l}{\emph{SiT-XL/2, ODE (Heun), CFG $w=1.5$}} \\
Baseline & \xmark & \xmark & 5.59 & 157.6 & 4.71 & 0.79 & 0.54 \\
+ Dispersive Loss$^\ddagger$ & \xmark & \xmark & 5.01 & 166.1 & 4.76 & 0.80 & 0.54 \\
+ \textbf{SPARE (Ours)} & \xmark & \xmark & \textbf{4.03} & \textbf{185.7} & \textbf{4.64} & \textbf{0.81} & \textbf{0.55} \\
\bottomrule
\end{tabular}}
\caption{Main results on ImageNet-1K $256\times256$ with SiT
backbones at 400K iterations. Enc./Par.: whether an external
encoder or extra trainable parameters are required (\ding{51} yes,
\ding{55} no). Competitor metrics are quoted from their papers;
$\ddagger$ marks our runs under the identical protocol; ``--'' means
not reported. Reported numbers from the Dispersive Loss paper
(32.05 on B/2 and 15.95 on XL/2 under ODE, 5.09 under CFG) are
comparable to our reproductions. \textbf{Bold} is best among
parameter-free methods, \underline{underline} best among
encoder-free methods. Methods are ordered by decreasing cost.}
\label{tab:imagenet-main}
\end{table}

\subsection{Main Results}
\label{sec:main_results}
\textbf{Comparison under the 400K protocol.}
Table~\ref{tab:imagenet-main} organizes existing regularizers along two costs, the source of the alignment target and the extra parameters required to consume it. REPA obtains the target from an external encoder and reduces FID by 12.40 on SiT-B/2 and 9.53 on SiT-XL/2, and SRA2 substitutes clean-data features, retaining roughly 60\% of that reduction; both keep a learnable head. Dispersive Loss removes all added parameters but discards the target as well. SPARE breaks this trade-off, holding a structured target at zero
added cost. Against Dispersive Loss at strictly equal cost, SPARE attains
lower FID in every block, by 1.32 to 2.39 without guidance and
0.98 with it. IS improves throughout, Precision is equal or better,
and sFID is ahead on XL/2 and marginally behind on B/2. Relative to REPA's quoted numbers, SPARE
recovers 54\% of the encoder-based reduction on B/2 and 37\% on
XL/2. Under classifier-free guidance the advantage persists, 4.03
against 5.01, with the largest IS and lowest sFID in the block.

\textbf{Composition with REPA.}
In Table~\ref{tab:imagenet-main}, SPARE alone remains behind REPA,
since only REPA draws on the semantics of a pretrained encoder.
Table~\ref{tab:repa_plugin} shows the two can simply be combined.
We add the affinity loss to REPA's recipe with nothing else
changed, so the affinity target stays the clean latent, never the
encoder's features. SiT-XL/2 FID improves from 7.90 to 7.49 at 400K
iterations, with IS, Precision, and Recall also ahead, and the
advantage holds at every checkpoint from 100K to 400K
(Figure~\ref{fig:efficiency}). At 1M iterations under CFG, FID
improves from 1.96 to 1.90 with sFID also ahead. Recall rises from
0.59 to 0.61 while IS dips slightly, a shift toward broader
coverage with Precision within 0.01. SPARE therefore adds to
encoder-based alignment rather than substituting for it.

\textbf{Text-to-image composition.}
Table~\ref{tab:t2i} extends the composition beyond class
conditioning, pairing MMDiT+REPA on MS-COCO with and without the
affinity loss. Under the 50-step SDE sampling typical of
text-to-image practice, SPARE lowers FID from 6.03 to 4.93, so the latent's affinity structure stays informative under text
conditioning.
 
\textbf{Training cost.}
Table~\ref{tab:cost} quantifies the cost annotations of
Table~\ref{tab:imagenet-main}. REPA adds 94.1M parameters and 21\%
forward compute through its encoder and head. DiverseDiT consults
no external features, yet its cross-layer repulsion costs 37M
parameters, 5.5~GB of memory, and 19\% of throughput, the largest
footprint after REPA despite holding no target. Dispersive Loss
and SPARE add neither parameters nor forward compute but differ in
memory, 1.1~GB for batch-wide negatives against 0.08~GB for the
affinity target. SPARE alone trains at essentially the footprint
of the unmodified backbone.

\begin{figure*}[!t]
\centering
\includegraphics[width=0.87\textwidth]{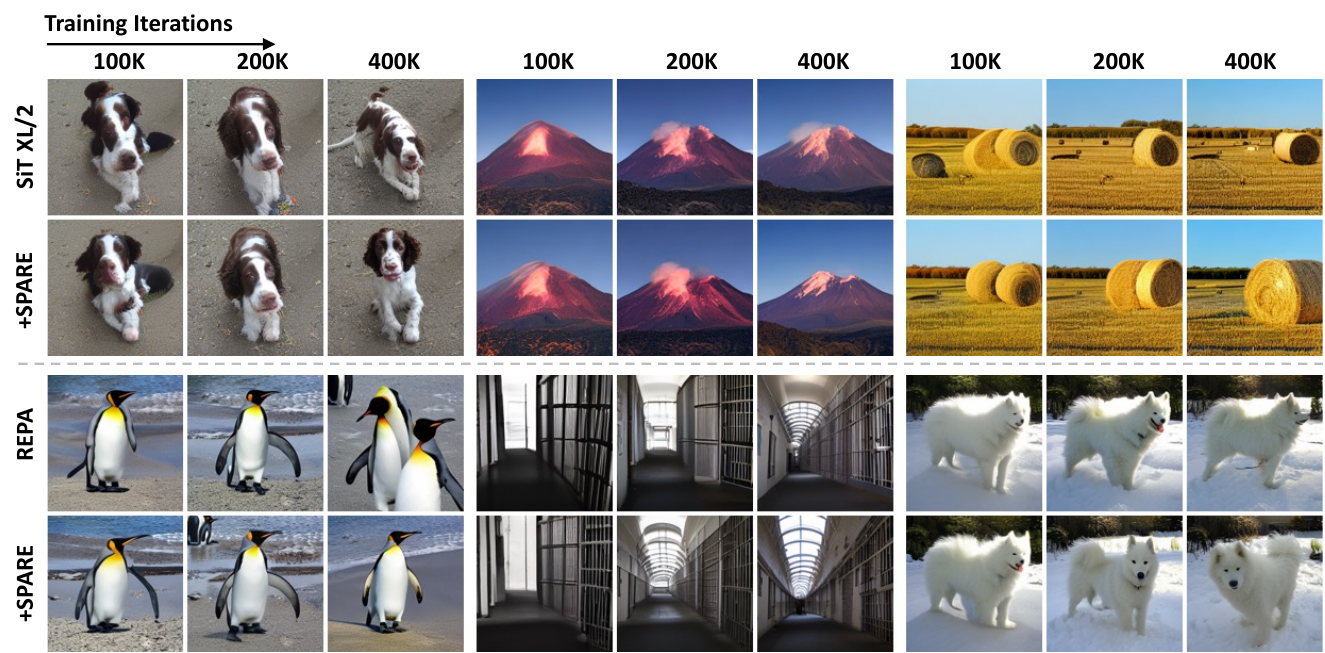}
\caption{SiT-XL/2 generations at 100K, 200K, and 400K iterations
(CFG $w=4.0$, SDE, 250 steps), with class label and initial noise
fixed per column group. \textbf{Top}: baseline versus $+$SPARE; \textbf{Bottom}: REPA versus $+$SPARE. SPARE-trained models reach stable object structure earlier.}
\label{fig:qualitative}
\end{figure*}

\textbf{Qualitative Results.}
\label{sec:qualitative}
Figure~\ref{fig:qualitative} compares SiT-XL/2 generations across
training. SPARE-trained models settle on coherent object layout
earlier, with the 200K geometry already close to the 400K result,
consistent with the FID trajectories in
Fig.~\ref{fig:efficiency}. More samples are in the supplementary
material.

\begin{table}[t]
\centering
{\small
\setlength{\tabcolsep}{3pt}
\begin{tabular}{llccccc}
\toprule
Method & Iter. & FID$\downarrow$ & sFID$\downarrow$ & IS$\uparrow$ & Prec.$\uparrow$ & Rec.$\uparrow$ \\
\midrule
\multicolumn{7}{l}{\textit{SiT-B/2, SDE (Euler--Maruyama), w/o CFG}} \\
REPA        & 400K & 24.40 & 6.40 & 59.9 & 0.59 & 0.65 \\
\ + SPARE   & 400K & \textbf{23.60} & \textbf{6.39} & \textbf{62.4} & 0.59 & 0.65 \\
\midrule
\multicolumn{7}{l}{\textit{SiT-XL/2, SDE (Euler--Maruyama), w/o CFG}} \\
REPA        
            & 400K & 7.90  & 5.06 & 122.6 & 0.70 & 0.65 \\
\ + SPARE   
            & 400K & \textbf{7.49}  & \textbf{5.05} & \textbf{129.7} & \textbf{0.71} & \textbf{0.66} \\
\midrule
\multicolumn{7}{l}{\textit{SiT-XL/2, SDE, w/ CFG ($w=1.35$), 1M iter.}} \\
REPA        & 1M   & 1.96 & 4.53 & \textbf{264.0} & \textbf{0.82} & 0.59 \\
\ + SPARE   & 1M   & \textbf{1.90} & \textbf{4.49} & 257.4 & 0.81 & \textbf{0.61} \\
\bottomrule
\end{tabular}}
\caption{SPARE as a plugin on REPA on ImageNet 256$\times$256. REPA
rows are quoted from the original paper; +SPARE rows add only the
affinity loss to REPA's recipe. The 1M CFG pair is our run
for both rows at the same guidance scale. \textbf{Bold} is the
better value within each matched pair; ties are unmarked.}
\label{tab:repa_plugin}
\end{table}

\begin{table}[t]
\centering
{\small
\setlength{\tabcolsep}{6pt}
\begin{tabular}{lc}
\toprule
Method & FID$\downarrow$ \\
\midrule
MMDiT + REPA & 6.03 \\
\ + SPARE    & \textbf{4.93} \\
\bottomrule
\end{tabular}}
\caption{Text-to-image composition on MS-COCO 256$\times$256.
Both rows are our runs of MMDiT under REPA's recipe; +SPARE adds
only the affinity loss.}
\label{tab:t2i}
\end{table}

\begin{table}[!t]
\centering

{\small
\setlength{\tabcolsep}{2.5pt}
\renewcommand{\arraystretch}{1.05}
\begin{tabular}{@{}lcccc@{}}
\toprule
\textbf{Method} & \textbf{Params (M)} & \textbf{$\Delta$FLOPs (G)} & \textbf{it/s} & \textbf{$\Delta$Mem (GB)} \\
\midrule
SiT-XL/2 & 0 $+$ 0 & --      & 3.36 & --      \\
\midrule
REPA             & 86 $+$ 8.1 & $+$24.1 & 3.09 & $+$0.67 \\
DiverseDiT       & 0 $+$ 37   & $+$9.5  & 2.72 & $+$5.52 \\
SRA2             & 0 $+$ 18.6 & $+$4.8  & 3.27 & $+$0.57 \\
Disp.\ Loss      & 0 $+$ 0    & 0       & \textbf{3.35} & $+$1.14 \\
\textbf{SPARE (Ours)} & 0 $+$ 0 & 0     & 3.33 & $+$\textbf{0.08} \\
\bottomrule
\end{tabular}
}
\caption{Training cost on SiT-XL/2 (batch 32, single H20). Params counts the external encoder $+$ projection head; $\Delta$FLOPs and $\Delta$Mem are per-forward FLOPs and peak memory over the baseline (114.4 GFLOPs, 22.56 GB).}
\label{tab:cost}
\end{table}

\subsection{Ablation Studies}
\label{sec:ablations}
\begin{table}[!t]

\centering
{\small
\setlength{\tabcolsep}{2.5pt}
\renewcommand{\arraystretch}{1.1}
\begin{tabular}{lccccc}
\toprule
& \textbf{FID}$\downarrow$ & \textbf{IS}$\uparrow$ & \textbf{sFID}$\downarrow$ & \textbf{Prec.}$\uparrow$ & \textbf{Rec.}$\uparrow$ \\
\midrule
baseline     & 36.49 & 40.60 & 6.47 & 0.51 & 0.65 \\
\midrule
\multicolumn{6}{l}{\emph{(a) Timestep interval} ($\ell=2, \tau = 1.0$)} \\
\quad $t \in [0,1]^\dagger$      & 32.41 & \textbf{45.24} & 6.27 & 0.53 & \textbf{0.65} \\
\quad $t \in [0,0.5]$            & 32.80 & 44.25 & 6.24 & 0.53 & \textbf{0.65} \\
\quad $t \in [0.5,1]$            & \textbf{32.30} & 44.79 & \textbf{6.21} & \textbf{0.54} & 0.64 \\
\midrule
\multicolumn{6}{l}{\emph{(b) Injection block} $\ell$ ($t \in [0,1], \tau = 1.0$)} \\
\quad $\ell = 2$                 & 32.41 & 45.24 & \textbf{6.27} & 0.53 & \textbf{0.65} \\
\quad $\ell = 3^\dagger$         & \textbf{31.94} & \textbf{45.65} & 6.32 & \textbf{0.54} & \textbf{0.65} \\
\quad $\ell = 4$                 & 32.52 & 44.68 & 6.40 & 0.53 & \textbf{0.65} \\
\quad $\ell = 8$                 & 36.52 & 39.90 & 6.55 & 0.51 & \textbf{0.65} \\
\midrule
\multicolumn{6}{l}{\emph{(c) Temperature} $\tau$ ($t \in [0,1], \ell=3$)} \\
\quad $\tau = 0.05$              & 32.16 & 45.47 & 6.38 & 0.54 & 0.64 \\
\quad $\tau = 0.5^\dagger$       & \textbf{31.85} & 45.60 & \textbf{6.30} & \textbf{0.54} & \textbf{0.65} \\
\quad $\tau = 1.0$               & 31.94 & \textbf{45.65} & 6.32 & \textbf{0.54} & \textbf{0.65} \\
\quad $\tau = 2.0$               & 33.53 & 43.58 & 6.32 & 0.53 & 0.64 \\
\midrule
\multicolumn{6}{l}{\emph{(d) Cross-image target} ($t \in [0,1], \ell=3$)} \\
\quad Intra-only ($\tau{=}1$)               & 31.94 & 45.65 & \textbf{6.32} & \textbf{0.54} & \textbf{0.65} \\
\quad + zero (repel, $\tau{=}1$)            & 34.73 & 42.71 & 6.71 & 0.52 & 0.64 \\
\quad + VAE (align, $\tau{=}1$)             & 31.85 & 45.71 & 6.33 & \textbf{0.54} & 0.64 \\
\quad + VAE (align, $\tau{=}0.5$)$^\dagger$ & \textbf{31.59} & \textbf{46.58} & 6.42 & \textbf{0.54} & 0.64 \\
\midrule
\multicolumn{6}{l}{\emph{(e) Loss weight} $\lambda$ ($t \in [0,1]$, $\ell=3$,
$\tau=0.5$)} \\
\quad $\lambda = 0.5$            & \textbf{31.53} & \textbf{46.93} & \textbf{6.31} & \textbf{0.54} & \textbf{0.65} \\ 
\quad $\lambda = 1.0^\dagger$    & 31.59 & 46.58 & 6.42 & \textbf{0.54} & 0.64 \\
\quad $\lambda = 2.0$            & 32.12 & 46.13 & 6.54 & \textbf{0.54} & 0.64 \\
\bottomrule
\end{tabular}}
\caption{Ablation studies on SiT-B/2 (400K iterations, ODE Heun sampler,
250 steps, no CFG). $\dagger$ marks the setting used in the main
experiments. Panels (a)--(c) use the intra-image objective to isolate
each factor. In panel (d), the two cross-image variants share the candidate
set and joint softmax and differ only in the inter target (zero or
the VAE latent), while intra-only drops the inter candidates.}
\label{tab:ablation}
\end{table}

\begin{figure}[!t]
\centering
\includegraphics[width=\linewidth]{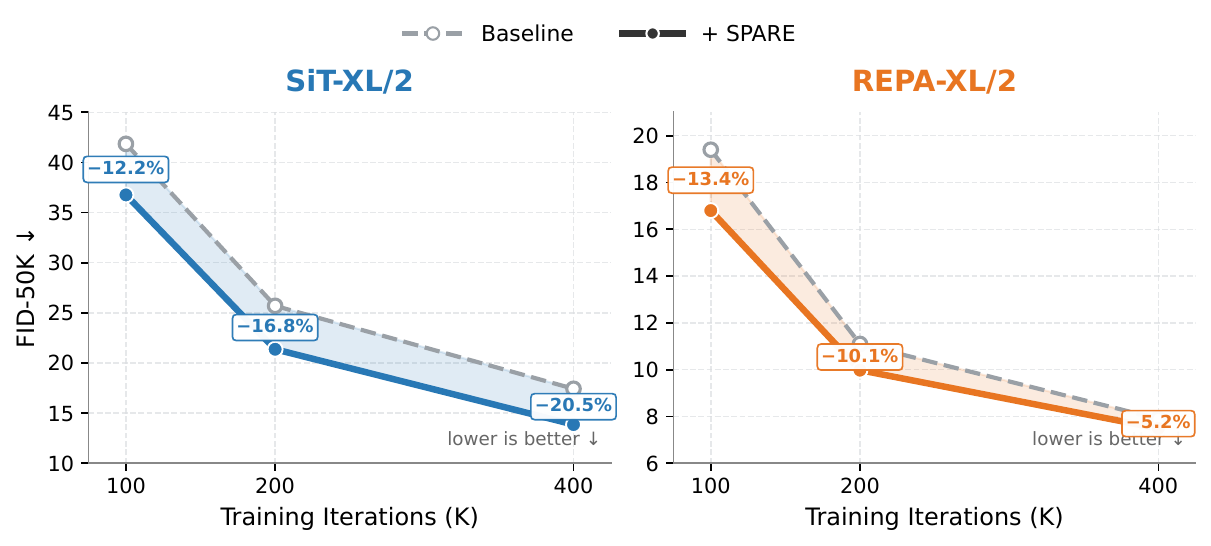}
\caption{FID-50K across training for the baseline (left) and REPA
(right) pairings on SiT-XL/2, without CFG. Annotations give the
relative reduction at each checkpoint. SPARE lowers the trajectory, and the gap holds as training proceeds.}
\label{fig:efficiency}
\end{figure}

\textbf{Training efficiency.}
Figure~\ref{fig:efficiency} plots the FID trajectories underlying Tables~\ref{tab:imagenet-main} and~\ref{tab:repa_plugin}. SPARE lowers the curve at every checkpoint, and the baseline needs its full 400K iterations to reach a FID that SPARE attains well earlier. The gain persists along REPA's stronger trajectory. 

\textbf{Timestep interval.}
Restricting the calibration to either half of the noise schedule
changes FID by at most 0.5 (Table~\ref{tab:ablation}a). The
clean-latent target is informative at every noise level, so the
full interval $t \in [0,1]$ requires no scheduling.

\textbf{Injection block.}
The choices $\ell \in \{2,3,4\}$ fall within 0.6 FID of one
another, with $\ell=3$ best (Table~\ref{tab:ablation}b). At
$\ell=8$ the gain vanishes and FID returns to the baseline (36.52
vs.\ 36.49), indicating that later representations have specialized
beyond the latent's spatial organization.

\textbf{Temperature.}
All of $\tau \in [0.05, 1]$ fall within 0.31 FID
(Table~\ref{tab:ablation}c). Only $\tau = 2$, which flattens the
target toward uniform, degrades noticeably, consistent with the
structure rather than the sharpness of the target carrying the
signal.

\textbf{Cross-image target.}
Table~\ref{tab:ablation}d isolates whether cross-image relations
should be calibrated or repelled. The two cross-image variants keep
the candidate set and joint softmax of Eq.~\eqref{eq:spare_loss}
and differ only in the target mass on the inter candidates, while
intra-only removes the inter candidates altogether. Setting it to
zero leaves them only in the student's partition function, so
minimization repels every cross-image pair, a parameter-free analog
of Dispersive Loss. This is harmful, degrading FID by 2.79 over
intra-only, as pushing batch-mates apart destroys affinities the
data marks as high. Assigning the clean latent's affinity instead
improves over intra-only at zero cost. The 2.88 FID swing between
repelling and aligning under identical machinery is the paper's
claim in miniature. What matters is not whether cross-image
relations enter the loss, but whether they enter as a target or as
a casualty.
 
\textbf{Loss weight.}
SPARE is stable in $\lambda$ (Table~\ref{tab:ablation}e). Both
$\lambda=0.5$ and $\lambda=1$ improve over the baseline by at
least 4.9 FID, and $\lambda=2$ costs only 0.5. We keep
$\lambda=1$ throughout, transferred unchanged to XL/2.
 
\section{Conclusion}
We presented SPARE, a parameter-free regularizer that calibrates
the pairwise affinities of intermediate diffusion transformer
tokens to those of the clean VAE latents, within each image and at
matched positions across the batch. Under the 400K protocol on
ImageNet $256 \times 256$, SPARE attains the lowest FID among
parameter-free regularizers in every tested setting, recovers 37
to 54\% of REPA's FID reduction at negligible cost, and improves
over REPA when added to it. The ablations place the effect in the
target rather than the machinery, since the same candidate set
helps when cross-image affinities are matched and hurts when they
are repelled.

Encoder-based alignment retains an edge at this budget, and the
composition results indicate the two signals are complementary
rather than interchangeable. Most findings use SiT backbones on
ImageNet $256 \times 256$ at 400K iterations, with single runs at
1M and on MS-COCO text-to-image; higher resolutions and larger
text-to-image scale remain open. Affinity calibration requires
nothing beyond the latents themselves, so extending the target to
richer structure, and the candidate set beyond matched positions,
are natural directions.

\bibliography{aaai2027}


\end{document}